\documentclass[pdflatex,sn-mathphys]{sn-jnl}

\usepackage{color}
\usepackage{xspace}

\makeatletter
\DeclareRobustCommand\onedot{\futurelet\@let@token\@onedot}
\def\@onedot{\ifx\@let@token.\else.\null\fi\xspace}

\def\etal{\emph{et al}\onedot}
\makeatother

\jyear{2022}%

\theoremstyle{thmstyleone}%
\theoremstyle{thmstyletwo}%

\theoremstyle{thmstylethree}%

\begin{document}

\title[MPF6D: Masked Pyramid Fusion 6D Pose Estimation]{MPF6D: Masked Pyramid Fusion 6D Pose Estimation}


\author*[1]{\fnm{Nuno} \sur{Pereira}}\email{nuno.pereira@ubi.pt}

\author[1]{\fnm{Lu\'{\i}s} \sur{A. Alexandre}}\email{luis.alexandre@ubi.pt}


\affil*[1]{\orgdiv{}, \orgname{Departamento de Inform\'{a}tica and NOVA LINCS, Universidade da Beira Interior}, \orgaddress{\street{Rua Marqu\^{e}s de \'{A}vila e Bolama, Convento de Santo Ant\'{o}nio}, \city{Covilhã}, \postcode{6200-001}, \state{Castelo Branco}, \country{Portugal}}}




\abstract{
Object pose estimation has multiple important applications, such as robotic grasping and augmented reality. We present a new method to estimate the 6D pose of objects that improves upon the accuracy of current proposals and can still be used in real-time. Our method uses RGB-D data as input to segment objects and estimate their pose. It uses a neural network with multiple heads to identify the objects in the scene, generate the appropriate masks and estimate the values of the translation vectors and the quaternion that represents the objects' rotation. These heads leverage a pyramid architecture used during feature extraction and feature fusion. We conduct an empirical evaluation using the two most common datasets in the area, and compare against state-of-the-art approaches, illustrating the capabilities of MPF6D. Our method can be used in real-time with its low inference time and high accuracy.
}




\maketitle

\section{Introduction}\label{sec1}
In the context of Industry 4.0, robotic systems require adaptation to handle unconstrained pick and place tasks, human-robot interaction and collaboration, and autonomous robot movement. These environments and tasks are dependent on methods that perform object detection, object localization, object segmentation, and object pose estimation. To have accurate robotic manipulation, unconstrained pick and place, and scene understanding, accurate object detection and pose estimation methods are needed. These methods are used in other contexts like augmented reality, for example, where badly placed objects into the real-world break the experience of augmented reality. Another application example is the use of augmented reality in the industry to train new and competent workers where virtual objects need to be placed in the correct positions to look like real objects or simulate their placement in the correct positions.

6D pose estimation is a hard problem to tackle due to the possible scene cluttering, illumination variability, object truncations, and different shapes, sizes, textures, and similarities between objects.

We present a new 6D pose estimation method that has low pose estimation error and can be used in real-time. Our method is a complete pipeline that can detect, segment, and estimate the 6D pose of known objects presented in the scene.

Some methods, like MaskedFusion\cite{maskedfusion} or DenseFusion \cite{densefusion}, have two individual traning steps, one to train the detection and segmentation neural network and another for training the neural network responsible for estimating the objects' pose. This last neural networks needs to have the previous step trained. MPF6D, on the other hand is trained as a single method. Besides having a simpler and faster training MPF6D achieves higher accuracy due to the new architecture that leverages a pyramid neural network to use object features extracted at different resolutions.

The main contributions are:
\begin{itemize}
    \item A new high accuracy 6D pose estimation method;
	\item A simpler pipeline for 6D pose estimation that can be trained as a single method, this facilitates the use of MPF6D in different scenarios, datasets or type of objects;
	\item A fast method that can be used in real-time, achieving 8 frames per second;
	\item An experimental evaluation using the standard datasets used for assessing 6D pose estimation methods including a comparison against the current best methods in the area.
\end{itemize}

\begin{figure}[thpb]
   \centering
   \includegraphics[width=\columnwidth]{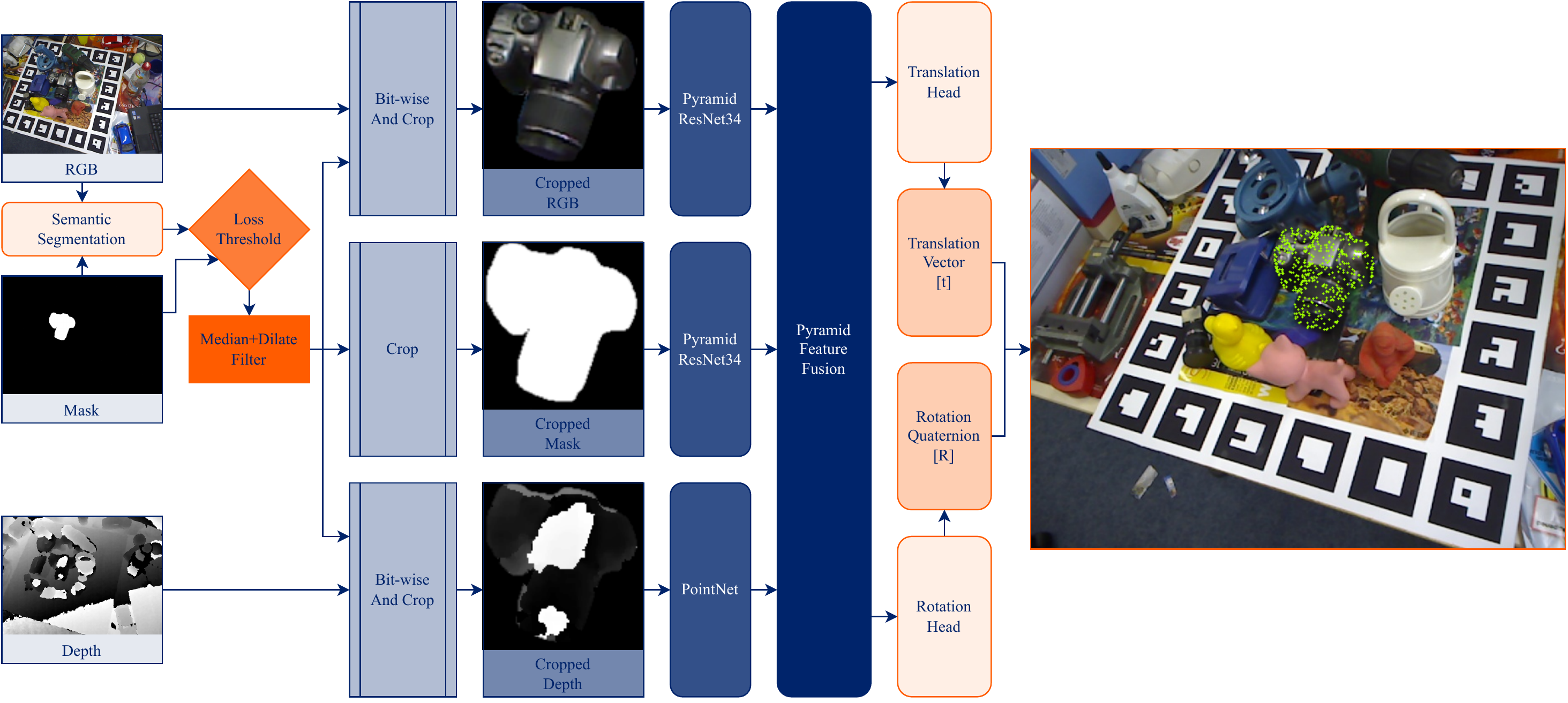}
   \caption{Representation of the data-flow thought MPF6D without the Pose Refinement optional step. Note that this flow is for the training phase. At inference time, the mask is generated by our method using the semantic segmentation head.}
   \label{fig:Overview}
\end{figure}

\section{Related Work}
In this section, we present the most relevant literature related to object 6D pose estimation. First, we introduce methods for semantic segmentation, and then we present methods that estimate the object's pose.

\subsection{Semantic Segmentation}
One of the most notable methods for semantic segmentation is the U-Net \cite{unet} convolutional neural network that was originally developed for segmenting biomedical images. Its name comes from the architectures U letter-like shape. The U-Net architecture is composed of two parts, the left part is the contracting path and the right part is the expansive path. The contracting path's purpose is to capture context while the purpose of the expansive path is to assign the classes to the pixels based on the context. SegNet \cite{segnet} has a similar architecture to U-Net where the second half consists of the same structure as in the first half but hierarchically symmetric.
SegNet uses a fully convolutional neural network based on the VGG-16 \cite{VGG16} convolutional layers.
FastFCN \cite{fastfcn} architecture uses a Joint Pyramid Upsampling module instead of using dilated convolutions to assign classes to the pixels, since those consume more memory and time during training. FastFCN has a fully connected network as its backbone and the Joint Pyramid Upsampling is used to upsample the features and label the pixels. This pyramid upsamples the low-resolution feature maps into high-resolution feature maps.
The Pyramid Scene Parsing Network (PSPNet) \cite{pspnet}, uses global contextual information for semantic segmentation. The authors introduced a Pyramid Pooling Module after the last layers of a fully convolutional neural network, that is based on ResNet-18, and the feature maps obtained from the fully convolutional neural network are pooled using four different scales corresponding to four different pyramid levels. The polled feature maps are then convoluted using a 1×1 convolution layer to reduce the feature maps dimension. These outputs of each convolution are then upsampled and concatenated with the initial feature maps that were extracted from the fully convolution neural network. This concatenation provides the local and global contextual information of the image. After the concatenation, the authors use another convolution layer to generate the final pixel-wise predictions.
The PSPNet objective is to observe the whole feature map in sub-regions with different locations using the pyramid pooling module.
The Gated-SCNN \cite{gated-scnn} architecture consists of a two-stream convolutional neural network architecture. The first stream branch is used to process image shape information, and the second is used to process boundary information. A gating method is used in the intermediate layers of each branch to connect features from both branches. In the end, it fuses all the features from both branches and predicts the semantic segmentation masks. One challenge with using deep fully convolutional neural networks on images for segmentation tasks is that input feature maps become smaller while traversing through the convolutional and pooling layers of the network. This causes loss of information and results in an output where predictions are of low resolution and object boundaries are fuzzy. DeepLab models \cite{deeplabv3} address this challenge by using Atrous convolutions and Atrous Spatial Pyramid Pooling modules. The DeepLab architecture has evolved over several generations: DeepLabV1 uses Atrous Convolution and Fully Connected Conditional Random Field to control the resolution at which image features are computed. DeepLabV2 uses Atrous Spatial Pyramid Pooling to consider objects at different scales and segments with improved accuracy. Finally, DeepLabV3, apart from using Atrous Convolution, also uses an improved Atrous Spatial Pyramid Pooling module by including batch normalization and image-level features, and it does not use the Conditional Random Field as in previous versions.

\begin{figure*}[tb]
   \centering
   \includegraphics[width=.9\textwidth]{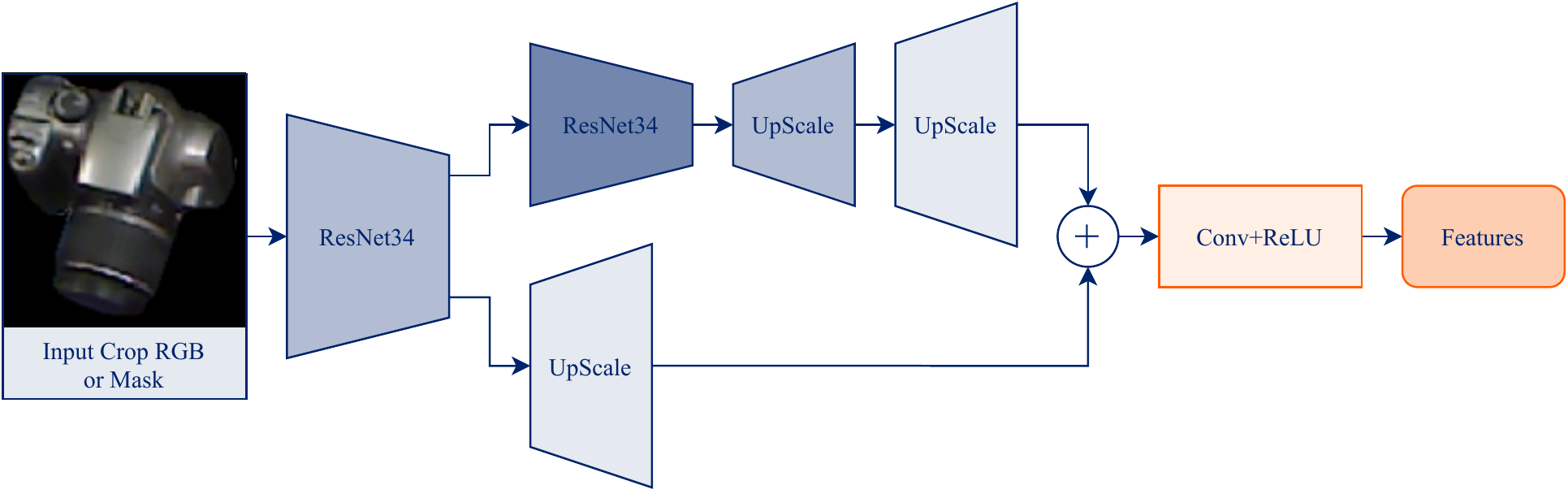}
   \caption{In-depth representation of Pyramid ResNet34. This system is used with both RGB and Mask images to extract features, with only one change between them: the first layer for the mask receives only one channel instead of three.}
   \label{fig:PyramidResNet34}
\end{figure*}

\subsection{6D Pose Estimation}
The methods for 6D Pose Estimation can be divided, based on the type of input that they use, into three different categories, RGB, Point Cloud, and RGB-D methods. Methods like \cite{pnpex1, pnpex2, pnpex3, pnpex4, pvnet} that use RGB images as input, rely on the detection and matching of keypoints from the objects with keypoints from the object’s 3D render and then use the PnP \cite{pnp} algorithm to estimate the 6D pose.
In this category, there are other methods, such as \cite{localrgbd} that do template matching with cropped patches from the object presented in the image and approximate the 3D model of the objects to the cropped patches to estimate the object's pose.
Point cloud methods \cite{pvfh, frustum, pointnet, vfh, voxelnet} rely on descriptors to extract object features, and match the extracted features with features acquired from known poses.
RGB-D methods \cite{ssd6d, localrgbd, liunified, densefusion, posecnn, pointfusion, pvn3d} regress the 6D poses directly regressed from the input data. Usually, these methods have a pose refinement phase using the Iterative Closest Point algorithm where the depth data is mostly taken into account. In this category, fewer methods \cite{maskedfusion, densefusion, pvn3d} use the RGB and depth as input data to achieve better pose accuracies and also use refinement phases to achieve higher accuracy.
Tejani \etal \cite{tejani} follow a local approach where small RGB-D patches vote for object pose hypotheses in a 6D space.
Kehl \etal \cite{localrgbd} also follow a local approach but they use a convolutional auto-encoder (CAE) to encode each patch of the object to later match the obtained features in the bottleneck of the CAE with a code-book of features learned during the train and use the code-book matches to predict the 6D pose of the object.
Although such methods are not taking global context into account, they proved to be robust to occlusion and the presence of noise artifacts since they infer the object pose using only small patches of the image.
SSD-6D \cite{ssd6d} uses an RGB image that is processed by the NN to output localized 2D detections with bounding boxes, classifies the bounding boxes into discrete bins of Euler angles and subsequently estimates the object's 6D pose.
This method is in the RGB-D category because after the first estimation, and with the availability of the depth information, the 6D poses can be further refined.
PoseCNN \cite{posecnn} uses a new loss function that is robust to object symmetry to directly regress the object rotation.
It uses a Hough voting approach to obtain the 3D center of the object to estimate its translation.
Using ICP on the refinement phase of SSD-6D and PoseCNN makes their 6D pose estimation more accurate.
Li \etal \cite{liunified} formulate a discriminative representation of 6-D pose that enables predictions of both rotation and translation by a single forward pass of a convolutional neural network, and it can be used with many object categories.
DenseFusion \cite{densefusion} extract features from RGB images and depth data with different fully convolutional neural network.
After the extraction, it fuses the depth and RGB features while retaining the input's space geometric structure.
DenseFusion is similar to PointFusion \cite{pointfusion}, as it also estimates the 6D pose while keeping the geometric structure and appearance information of the object, to later fuse this information in a heterogeneous architecture.
MaskedFusion \cite{maskedfusion} is a pipeline divided into 3 sub-tasks that combined can solve the task of object 6D pose estimation.
In the first sub-task, the detection and segmentation for each object in the scene occur.
The neural network presented in the first sub-task classifies each pixel of the RGB image captured and predicts the mask and the location for each object in the scene. After the output of the neural network, filters are applied to the mask and then a \textit{bit-wise AND} operation is used on the original RGB and depth images to crop the intended object.
In the second sub-task, with the masks obtained from sub-task 1 for each object and the RGB-D data, it is possible to estimate the object 6D pose.
For each type of input data, the method has different neural networks to extract features.
After all the features are extracted they are combined and another neural network is used to extract the most meaningful features and regress the estimated 6D pose of the object.
After this sub-task, the 6D pose is estimated, but it is also possible to do pose refinement using another neural network.
PVN3D \cite{pvn3d}, uses DenseFusion as a backbone to extract features of the object and then uses a shared MLP to estimate the object keypoints and segment each object. A clustering algorithm is then used to find the different points of the object. In the end, a least-squares fitting algorithm is used to estimate the 6D pose of the object.

Most of the related methods rely on an object detector or object segmentation method to pre-process the scene and then crop each object to then estimate its 6D pose.
Our method, MPF6D, uses RGB-D data and extracts features from both data, RGB and depth images, so it is categorized in the RGB-D category. MPF6D, does not need a pre-processing method to detect the object. It segments the objects in one neural network head while other heads estimate the translation of the object and its rotation.

\begin{figure*}[tb]
   \centering
   \includegraphics[width=.9\textwidth]{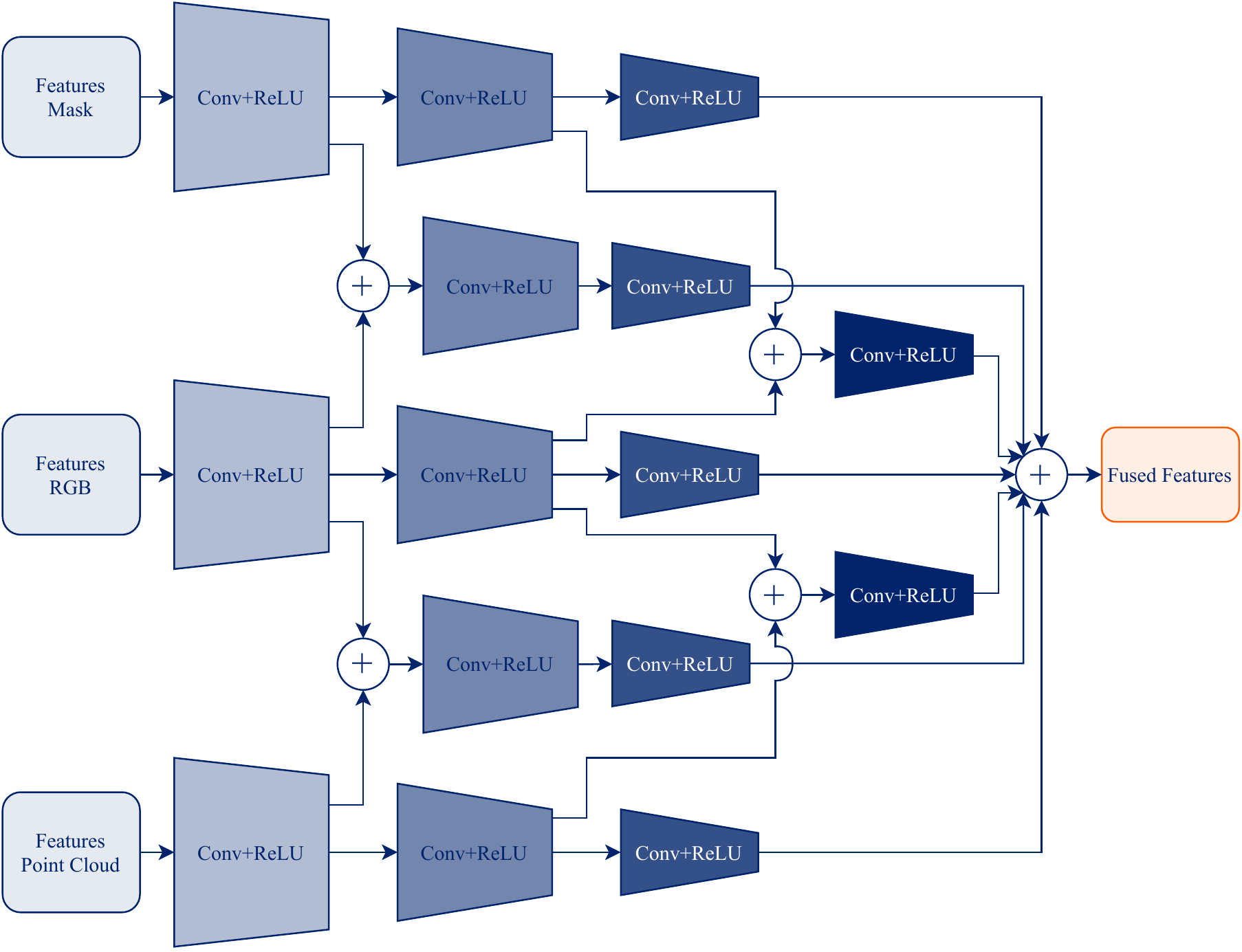}
   \caption{Pyramid Fusion is the architecture that fuses extracted features from the different data types (RGB, Mask, Depth/Point Cloud).}
   \label{fig:PyramidFusion}
\end{figure*}

\section{MPF6D}
Our method has influences from pyramid neural networks, as we use this type of architecture during the feature extraction and feature fusion so that we can have multiple features that combine low-resolution features, semantically strong features, and high resolution features.
With this type of architecture, it was possible to create an accurate and fast method to estimate the 6D pose of objects.

Our architecture has five steps from the input data until we have the estimated pose. The five steps are Semantic Segmentation, Feature Extraction, Feature Fusion, Estimation Heads and Pose Refinement. All five steps are trained simultaneously.

In Fig. \ref{fig:Overview}, we have the overview diagram that represents the data flow through MPF6D, except the optional Pose Refinement step. The flow starts on the top of the image with the RGB, Mask, and Depth images as input. Note that this flow is for the training phase. At inference time, the mask is generated using the semantic segmentation head. Fig. \ref{fig:PyramidResNet34} represents the pyramid scheme ResNet34 (we named it Pyramid ResNet34) as well as the upscale processes in order to have features from different scales in a pixel-wise form. In Fig. \ref{fig:PyramidFusion}, we show the fusion of the multiple extracted features that are used in the estimation heads (Translation Head, Rotation Head).

In the first step, Semantic Segmentation, we used the DeepLabV3 \cite{deeplabv3} architecture as a head of our method to detect, classify and generate the mask of the known objects presented in the scene. This head is trained using the same loss function that was proposed in \cite{deeplabv3}. The training is done at the same time as the rest of the method. This was possible due to the integration of this head in our neural network. 
After the prediction of the mask we apply a technique introduced by the authors of MaskedFusion \cite{maskedfusion}: first, we use a median filter to smooth the mask image with a kernel size of $3\times3$, and then, we dilate the mask with a $5\times5$ kernel such that, if the mask has some minor boundary segmentation error, this operation helps to correct it or complete any misclassified pixel in the middle of the object.
Since the 6D pose estimation step of our neural network requires an isolated object as input, we use the mask of the classified object to crop that object, and for that, we needed to add a mechanism that would enable the rest of the neural network to be trained efficiently instead of waiting for the segmentation head to be accurate. This mechanism is only used in the initial training epochs. The mechanism consists in using the ground truth mask in the 6D pose estimation step while the semantic segmentation head does not achieve a low and stable loss. We use the ground truth masks until the threshold mechanism detects that the loss of the segmentation head is stabilizing. Then we start to use masks produced by the segmentation head into the next steps. We apply and analyze the loss threshold in the validation subset. We use a \textit{bit-wise AND} between the mask and the RGB image and the mask and depth image in order to only have the pixels that have the object present in it. Then we do a rectangular crop of the object from the resultant images. This enables us to have a tensor that has a smaller size than using the full image as input tensor, where most of the pixels were black due to our \textit{bit-wise AND} operation to remove the background.

For the Feature Extraction step of the RGB and mask data, we use a ResNet34 architecture in a pyramid-like architecture (Fig. \ref{fig:PyramidResNet34}) where we stacked two of the original ResNet34 architectures to have multi-scale features.
This ResNet34 has at the end upscale layers. This type of layer is similar to the ones presented in PSPNet \cite{pspnet}. They consist of convolutional layers mixed with upsampling layers, that enables us to assign features for each original pixel of the object (pixel-wise features). We use the features produced by the first ResNet34 as input to the next ResNet34 and we upscale these features to the original object size after the first ResNet34 and after the second ResNet34, then we fuse both of the upscaled features. The fusion is made with a concatenation of the features tensors and the output of two convolutional layers. These fused features for the RGB data and mask will be used in the pyramid feature fusion.

For the depth data, we convert the cropped depth image into a point cloud, and then we use a PointNet architecture to extract features from the generated point cloud.

The neural network responsible for the feature extraction generates features corresponding to each different data type. For the feature fusion step (Fig. \ref{fig:PyramidFusion}), we use a pyramid like architecture with the intent of having multi-dimensional features with different scales that are then upscaled to the original object image size. 
The pyramid feature fusion has three different resolutions.
With this type of technique, which was previously used during the feature extraction of RGB and mask data, we can have the most significant features of the object while keeping the original features that represent the object's size, geometry, and pixel-wise multi-feature. With all the features fused we can then use the neural network heads to estimate the different values for the position of the object and its rotation.
We use two regression heads, one for estimating the translation vector of the object, and the other to estimate the quaternion that corresponds to the rotation of the object. After having this preliminary 6D pose of the object, we could use other methods (ICP or DenseFusion refinement) to refine the 6D pose estimation. However, we improved the DenseFusion refinement neural network to use it in our refinement step. We choose to improve upon DenseFusion since their refinement neural network can be used during the inference time without using too much computation time. We added two extra layers to use the same type of pyramid architecture that we used before to maintain the original scale of the features and have deeper features.

To train MPF6D we use the following loss function (\ref{eq:loss}) where we calculate the error between $M$ randomly sampled points and the ground truth object pose:

\vspace{-.2em}
\begin{equation}
    \label{eq:loss}
    \mathcal{L}^{p}_{i} = \frac{1}{M} \sum_{j} \left \| ( Rx_j + t) - (\hat{R_i}x_j + \hat{t_i}) \right \|
\vspace{-.5em}
\end{equation}

where, $x_j$ denotes the $j^{th}$ point of the $M$ randomly selected 3D points from the object's 3D model, $p = [R\vert t]$ is the ground truth pose, $R$ is the rotation matrix of the object and $t$ is the translation vector.
The estimated pose from MPF6D is represented by $\hat{p}_i = [ \hat{R}_i\vert \hat{t}_i]$ where $\hat{R}$ denotes the predicted rotation and $\hat{t}$ the predicted translation.

All these techniques in conjunction enable us to have and accurate 6D pose estimation while keeping the inference time as low as possible to enable our method to be used in real-world applications.

\begin{table*}[tb]
\centering
\caption{Quantitative evaluation of 6D pose using the ADD (\ref{eq:add}) metric on the LineMOD dataset. Symmetric objects are presented in italic and were evaluated using ADD-S (\ref{eq:add-s}). Bold shows best results in a given row.}
\label{tab:LineMOD}
\resizebox{\textwidth}{!}{%
\begin{tabular}{r|c|ccccccc}
Objects         & \begin{tabular}[c]{@{}c@{}}MPF6D\\Avg (Stdev)\end{tabular} & MPF6D*        & PVN3D          & MaskedFusion   & DenseFusion    & PointFusion & SSD-6D+ICP   & Implicit+ICP \\  \hline
ape             & 98.9 (0.3)           & \textbf{99.2}  & 97.3           & 91.4           & 92.3           & 70.4        & 65.0           & 20.6           \\
bench vi.       & 99.7 (0.3)           & 99.5           & \textbf{99.7}  & 99.0           & 93.2           & 80.7        & 80.0           & 64.3           \\
camera          & 100.0 (0.0)           & \textbf{100.0} & 99.6           & 99.0           & 94.4           & 60.8        & 78.0           & 63.2           \\
can             & 98.9 (0.3)           & 99.2     & 99.5           & \textbf{100.0} & 93.1           & 61.1        & 86.0           & 76.1           \\
cat             & 99.7 (0.3)           & \textbf{100.0} & 99.8           & \textbf{100.0} & 96.5           & 79.1        & 70.0           & 72.0           \\
driller         & 99.8 (0.1)           & \textbf{99.8}  & 99.3           & 97.0           & 87.0           & 47.3        & 73.0           & 41.6           \\
duck            & 99.1 (0.2)           & \textbf{99.3}  & 98.2           & 92.5           & 92.3           & 63.0        & 66.0           & 32.4           \\
\textit{eggbox} & 100.0 (0.0)           & \textbf{100.0} & 99.8           & 99.1           & 99.8           & 99.9        & \textbf{100.0} & 98.6           \\
\textit{glue}   & 100.0 (0.0)           & \textbf{100.0} & \textbf{100.0} & \textbf{100.0} & \textbf{100.0} & 99.3        & \textbf{100.0} & 96.4           \\
hole p.         & 99.8 (0.1)           & 99.8     & 99.9           & \textbf{100.0} & 92.1           & 71.8        & 49.0           & 49.9           \\
iron            & 99.7 (0.0)           & \textbf{99.8}  & 99.7           & 96.9           & 97.0           & 83.2        & 78.0           & 63.1           \\
lamp            & 99.6 (0.3)           & \textbf{99.8}     & \textbf{99.8}  & 98.1           & 95.3           & 62.3        & 73.0           & 91.7           \\
phone           & 99.3 (0.2)           & 99.1           & \textbf{99.5}  & 99.0           & 92.8           & 78.8        & 79.0           & 71.0           \\  \hline
Average         & 99.6 (0.1)           & \textbf{99.7}  & 99.4           & 97.8           & 94.3           & 73.7        & 76.7           & 64.7      \\
\multicolumn{9}{c}{\small(*) Best of three repetitions.}
\end{tabular}
}
\end{table*}

\section{Experiments}
In these experiments, the results showed that our method is the best overall method to tackle the challenges presented in two datasets. We trained MPF6D three times from scratch, thus meaning all our weights start randomly all three times. We show the average results for the three runs that we trained and we also show the best run and compare these results with previous methods. For all training and inference experiments we use a desktop computer with SSD NVME, 64GB of RAM, an NVIDIA GeForce GTX 1080 Ti, and Intel Core i7-7700K CPU.

\subsection{Datasets}
To evaluate our method we use two benchmark datasets, LineMOD and YCB-Video. These datasets are widely used by previous state-of-the-art methods.
The LineMOD Dataset \cite{linemod} was captured with a Kinect, and it has the RGB and depth images automatically aligned. The dataset consists of 13 low-textured objects, annotated 6D poses, and object masks. The main challenges of this dataset are the cluttered scenes, texture-less objects, and illumination variations.

The YCB-Video Dataset \cite{originalycb} contains 21 YCB objects of varying shape and texture, and is composed of 92 RGB-D videos, each with a subset of the objects placed in the scene. It has 6D pose annotations and objects masks. The varying lighting conditions, image noise, and occlusions make this dataset a challenge.

\subsection{Evaluation Metrics}
As in previous works \cite{ssd6d, pvnet, densefusion, posecnn} for the LineMOD dataset we used the Average Distance of Model Points (ADD) (\ref{eq:add}) \cite{linemod} as metric of evaluation for non-symmetric objects and for the egg-box and glue we used the Average Closest Point Distance (ADD-S) (\ref{eq:add-s}) \cite{posecnn}.
\begin{equation}
    \label{eq:add}
    \textnormal{ADD} = \frac{1}{m} \sum_{x \in M} \left \| ( Rx + t) - (\hat{R}x + \hat{t}) \right \|
\end{equation}
\begin{equation}
    \label{eq:add-s}
    \textnormal{ADD-S} = \frac{1}{m} \sum_{x_1 \in M} \min_{x_2 \in M} \left \| ( Rx_1 + t) - (\hat{R}x_2 + \hat{t}) \right \|
\end{equation}

In the metrics (\ref{eq:add}) and (\ref{eq:add-s}), assuming the ground truth rotation $R$ and translation $t$ and the estimated rotation $\tilde{R}$ and translation $\tilde{t}$, the average distance calculates the mean of the pairwise distances between the 3D model points of the ground truth pose and the estimated pose. $M$ represents the set of 3D model points and $m$ is the number of points. For the symmetric objects, the matching between points is ambiguous for some poses, and that is why the ADD-S metric is used for symmetric objects.

In the YCB-Video evaluation we use the same metrics as in previous works \cite{posecnn} and \cite{densefusion}. So the evaluation as been done with the area under the ADD-S (\ref{eq:add-s}) curve (AUC).
Using these common metrics enable us to have a direct comparison between our method and previous methods.

\subsection{Results: LineMOD}
As previously stated, the main challenges of this dataset are the cluttered scenes, texture-less objects, and illumination variations. Even with these challenging conditions, our method had less pose error overall than all previous methods.

As presented in Table \ref{tab:LineMOD}, with a direct comparison of our method and PVN3D we improved by 0.3\%. This value might not be seen as much improvement but since our method, PVN3D, and even MaskedFusion are close to the zero error mark, all slight improvements are hard to get.
Extracting features from different resolutions enable a more accurate pose estimation, independent of the camera angle and object distance. MPF6D achieved the best accuracy in 9 objects out of 13, in its best execution. The values presented in the second column of Table \ref{tab:LineMOD} correspond to the best run. Each run was trained from scratch where all the weights were initialized randomly. The first column has the average values and the standard deviation that were obtained for all three runs. We can see that even the average of our three repetitions presents better results than any of the competing approaches.

\subsection{Results: YCB-Video}

In Table \ref{tab:YCB}, we present the quantitative evaluation using the area under the ADD-S (\ref{eq:add-s}) curve (AUC). Our method outperforms all previous methods. Comparing it with PVN3D we had 1.96\% more area under the curve. In the YCB-Video dataset, our method had the best accuracy for 19 objects out of 21. The two objects where we lose for the PVN3D are basically the same object (clamp) but with different sizes. As with LineMOD, the average of our three repetitions has better average result than any of the other methods.

\begin{figure*}[tb]
   \centering
   \includegraphics[width=\textwidth]{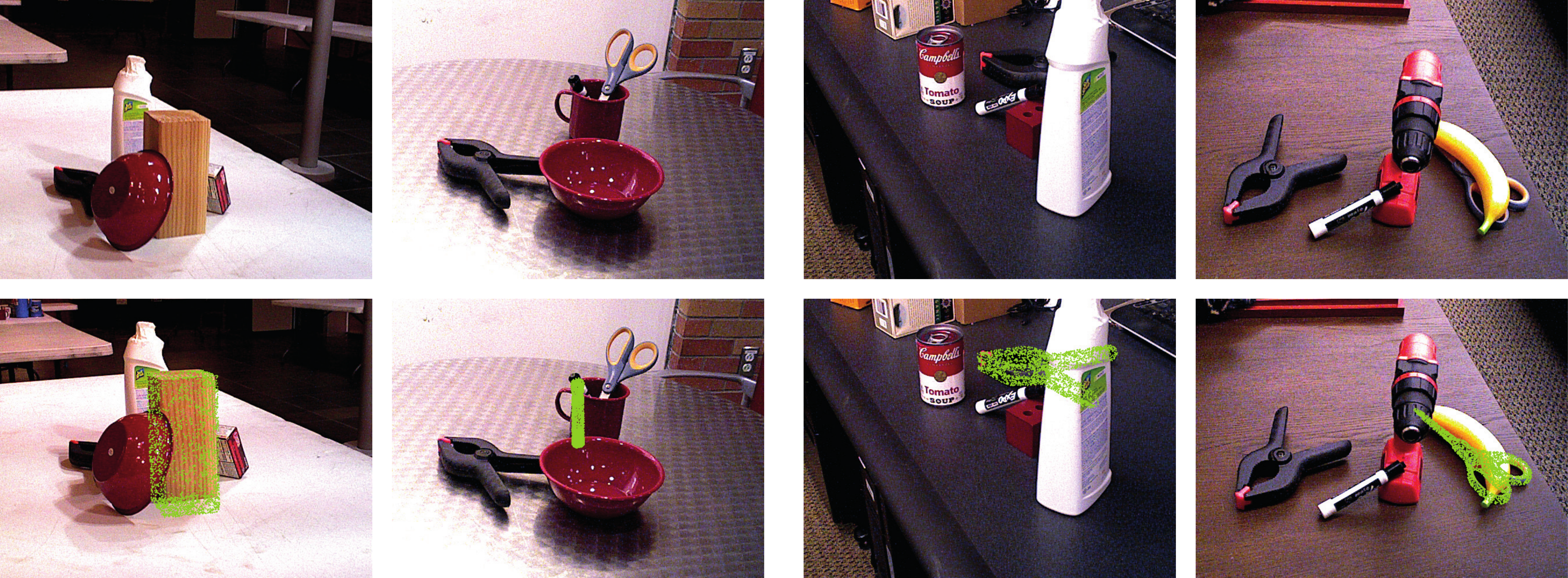}
   \caption{MPF6D object pose estimation examples. The green dots represent keypoints of the object pose estimation projected onto the RGB image. The top row contains the input RGB images and the bottom row the predicted object poses. The left two columns contain two examples of poor performance and the right two columns of good performance under heavy occlusion.}
   \label{fig:Results}
\end{figure*}

Figure \ref{fig:Results} shows two examples of poor performance of object pose estimation on the left and two good performance examples on the right. The two right examples are also good examples of the MPF6D handling object occlusions without losing accuracy.

\begin{table*}[tb]
\centering
\caption{Quantitative evaluation of 6D pose (area under the ADD-S (\ref{eq:add-s}) curve (AUC)) on the YCB-Video Dataset. Bold numbers are the best in a row.}
\label{tab:YCB}
\resizebox{\textwidth}{!}{%
\begin{tabular}{r|c|cccccc}
Objects                  & \begin{tabular}[c]{@{}c@{}}MPF6D\\Avg (Stdev)\end{tabular} & MPF6D*              & PVN3D+ICP    & MaskedFusion & DenseFusion & PointFusion & PoseCNN+ICP \\ \hline
002\_master\_chef\_can   & 99.69 (0.22)          & \textbf{99.44}       & 95.20          & 96.91        & 96.40       & 90.90       & 95.80       \\
003\_cracker\_box        & 99.60 (0.39)          & \textbf{99.75} & 94.40          & 96.55        & 95.50       & 80.50       & 92.70       \\
004\_sugar\_box          & 99.49 (0.21)          & \textbf{99.47}       & 97.90          & 98.81        & 97.50       & 90.40       & 98.20       \\
005\_tomato\_soup\_can   & 99.12 (0.28)          & \textbf{99.08}       & 95.90          & 95.64        & 94.60       & 91.90       & 94.50       \\
006\_mustard\_bottle     & 99.51 (0.37)          & \textbf{99.61} & 98.30          & 98.13        & 97.20       & 88.50       & 98.60       \\
007\_tuna\_fish\_can     & 99.50 (0.43)          & \textbf{99.52} & 96.70          & 97.31        & 96.60       & 93.80       & 97.10       \\
008\_pudding\_box        & 99.68 (0.40)          & \textbf{99.89} & 98.20          & 97.02        & 96.50       & 87.50       & 97.90       \\
009\_gelatin\_box        & 99.45 (0.24)          & \textbf{99.41}       & 98.80          & 98.69        & 98.10       & 95.00       & 98.80       \\
010\_potted\_meat\_can   & 97.47 (0.23)          & \textbf{97.64} & 93.80          & 94.57        & 91.30       & 86.40       & 92.70       \\
011\_banana              & 99.47 (0.41)          & \textbf{99.50} & 98.20          & 98.10        & 96.60       & 84.70       & 97.10       \\
019\_pitcher\_base       & 99.39 (0.46)          & \textbf{98.87}       & 97.60          & 97.06        & 97.10       & 85.50       & 97.80       \\
021\_bleach\_cleanser    & 99.02 (0.47)          & \textbf{98.99}       & 97.20          & 96.53        & 95.80       & 81.00       & 96.90       \\
024\_bowl                & 98.46 (0.20)          & \textbf{98.51} & 92.80          & 97.55        & 88.20       & 75.70       & 81.00       \\
025\_mug                 & 99.02 (0.45)          & \textbf{98.84}       & 97.70          & 97.48        & 97.10       & 94.20       & 95.00       \\
035\_power\_drill        & 99.15 (0.54)          & \textbf{99.54} & 97.10          & 97.26        & 96.00       & 71.50       & 98.20       \\
036\_wood\_block         & 95.81 (0.15)          & \textbf{95.98} & 91.10          & 95.42        & 89.70       & 68.10       & 87.60       \\
037\_scissors            & 97.13 (0.42)          & \textbf{97.10}       & 95.00          & 95.93        & 95.20       & 76.70       & 91.70       \\
040\_large\_marker       & 99.45 (0.40)          & \textbf{99.02}       & 98.10          & 97.55        & 97.50       & 87.90       & 97.20       \\
051\_large\_clamp        & 88.97 (5.09)          & 93.13          & \textbf{95.60} & 89.40        & 72.90       & 65.90       & 75.20       \\
052\_extra\_large\_clamp & 82.73 (5.93)          & 87.11          & \textbf{90.50} & 84.04        & 69.80       & 60.40       & 64.40       \\
061\_foam\_brick         & 98.85 (0.11)          & \textbf{98.83}       & 98.20          & 95.29        & 92.50       & 91.80       & 97.20       \\ \hline
Average                  & 97.66 (0.37)          & \textbf{98.06} & 96.10          & 95.96        & 93.20       & 83.90       & 93.00     \\
\multicolumn{8}{c}{\small(*) Best of three repetitions.}
\end{tabular}%
}
\end{table*}

\subsection{Inference}
\begin{table}[tb]
\centering
\caption{Quantitative inference time. The values presented in the table were measured in seconds.}
\label{tab:inference}
\resizebox{0.7\textwidth}{!}{%
\begin{tabular}{r|cccc}
Methods                           & Segmentation & 6D Pose & \begin{tabular}[c]{@{}c@{}}Pose\\Refinement\end{tabular} & Overall \\ \hline
DenseFusion                       & 0.03         & 0.02    & 0.01            & 0.06    \\
MPF6D                               & -            & -       & -               & 0.12    \\
PVN3D                             & -            & 0.02    & - (*)               & $>$ 0.17 (*)    \\
MaskedFusion                      & 0.2          & 0.01    & 0.002           & 0.212   \\
PoseCNN+ICP                    & 0.03         & 0.17    & 10.4            & 10.6   \\
\multicolumn{5}{c}{\small(*) Pose Refinement time not reported by the authors.}
\end{tabular}%
}

\end{table}

MPF6D can be used in real-time applications since it can infer the 6D pose of an object in $0.12$ seconds, so it can execute at 8 frames per second. This time was measured from the instant the data (RGB-D) was fed to the method until it produced the 6D pose estimation (translation vector and quaternion that has the rotation representation). We need an extra $0.02$ seconds to have the output as a translation vector and a rotation matrix.
PVN3D only reports the inference time on the LineMOD dataset, but in the YCB-Video the authors reported the results where they use the ICP to refine the 6D pose. Using the ICP algorithm improves the overall 6D pose estimation, but usually, this algorithm has high computation costs. It is possible to see that PoseCNN used the ICP refinement and just for the refinement it spent $10.4$ seconds to execute the ICP.

In Table \ref{tab:inference}, we present the quantitative comparison of values measured in seconds of methods that reported their inference times. The fastest method for inference is DenseFusion, but in terms of accuracy, we should compare the two best methods, ours and PVN3D, comparing these two we have $0.05$ seconds faster inference time than PVN3D even while using pose refinement.

\section{Ablation Studies}

\begin{table}[]
\centering
\caption{Ablation studies (using the ADD (\ref{eq:add})) on the LineMOD dataset. Ablation 1, removing the Pyramid ResNet34. Ablation 2, lower depth Pyramid Fusion. Ablation 3, removing Pyramid Fusion.}
\label{tab:ablationLineMOD}
\resizebox{0.7\columnwidth}{!}{%
\begin{tabular}{r|cccc}
Objects   & \begin{tabular}[c]{@{}c@{}}MPF6D\\Architecture\end{tabular} & \multicolumn{1}{c}{Ablation 1} & \multicolumn{1}{c}{Ablation 2} & \multicolumn{1}{c}{Ablation 3} \\  \hline
ape       & 90.3                                      & 75.2                                      & 79.5                                  & 75.5                                          \\
bench vi. & 90.5                                      & 73.1                                      & 79.8                                  & 75.3                                          \\
camera    & 96.7                                      & 78.7                                      & 85.4                                  & 82.0                                          \\
can       & 90.3                                      & 79.9                                      & 78.8                                  & 77.6                                          \\
cat       & 96.2                                      & 77.4                                      & 82.0                                  & 84.1                                          \\
driller   & 90.9                                      & 77.9                                      & 81.0                                  & 80.1                                          \\
duck      & 90.3                                      & 81.1                                      & 77.6                                  & 77.0                                          \\
\textit{eggbox}    & 96.9                                      & 78.4                                      & 87.0                                  & 82.4                                          \\
\textit{glue}      & 96.8                                      & 82.1                                      & 89.9                                  & 83.4                                          \\
hole p.   & 90.9                                      & 80.9                                      & 83.0                                  & 78.8                                          \\
iron      & 90.8                                      & 76.8                                      & 81.4                                  & 78.2                                          \\
lamp      & 90.8                                      & 76.0                                      & 83.4                                  & 75.2                                          \\
phone     & 90.2                                      & 77.9                                      & 82.2                                  & 77.9                                          \\ \hline
Average   & 92.4                                      & 78.1                                      & 82.4                                  & 79.0                                         
\end{tabular}%
}
\end{table}
For the ablation studies in our method, we performed three extra experiments in both datasets (LineMOD and YCB-Video).
For these experiments, we trained our method for 50 epochs in the LineMOD and YCB-Video datasets and we evaluated the inference output on the test subset of each dataset.
All the experiment results executed in the LineMOD dataset are shown in Table \ref{tab:ablationLineMOD} and the obtained results for the YCB-Video are shown in Table \ref{tab:ablationYCB}.
Ablation study one consists in testing the impact of removing the pyramid from the feature extraction step (Pyramid ResNet34), thus meaning that we only used a single ResNet34 and one upsample layer and then proceed to the fusion layers of the neural network.
With this experiment its possible to analyze the influence of our pyramid architecture on the MPF6D backbone feature extraction for the mask and RGB data.
Without the Pyramid ResNet34, the method obtained a 15\% increase in the error rate when compared to the original MPF6D architecture, thus meaning that having features extracted with multiple resolutions will improve the object 6D pose estimation.
The second ablation study focus on the impact of the Pyramid Fusion depth.
This study consists in removing one depth level of the Pyramid Fusion, thus enabling us to study if with a lower depth we could achieve the same results.
The obtained results had around 10\% more error overall.
The third ablation study evaluates the impact of removing the Pyramid Fusion from the original architecture. 
We replaced the Pyramid Fusion with just a simple concatenation of the multiple features, a convolution layer, and a ReLU activation function.
This third experiment showed that fusing multiple feature resolutions without using the pyramid approach, increases the overall error by 15\%

\begin{table}[]
\centering
\caption{Ablation studies (using area under the ADD-S (\ref{eq:add-s}) curve (AUC)) on the YCB-Video dataset. Ablation 1, removing the Pyramid ResNet34. Ablation 2, lower depth Pyramid Fusion. Ablation 3, removing Pyramid Fusion.}
\label{tab:ablationYCB}
\resizebox{\columnwidth}{!}{%
\begin{tabular}{r|cccc}
Objects                  & MPF6D Architecture & Ablation 1 & Ablation 2 & Ablation 3 \\ \hline
002\_master\_chef\_can   & 88.38 & 75.78                 & 79.01             & 73.98                     \\
003\_cracker\_box        & 88.66 & 75.79                 & 78.91             & 74.38                     \\
004\_sugar\_box          & 88.46 & 75.74                 & 79.56             & 75.09                     \\
005\_tomato\_soup\_can   & 88.32 & 75.62                 & 79.00             & 73.32                     \\
006\_mustard\_bottle     & 88.49 & 75.77                 & 79.04             & 73.91                     \\
007\_tuna\_fish\_can     & 88.11 & 74.94                 & 78.69             & 73.37                     \\
008\_pudding\_box        & 88.15 & 75.13                 & 78.37             & 73.17                     \\
009\_gelatin\_box        & 88.68 & 75.82                 & 79.17             & 74.45                     \\
010\_potted\_meat\_can   & 85.91 & 73.32                 & 76.34             & 72.53                     \\
011\_banana              & 88.83 & 76.37                 & 80.37             & 74.22                     \\
019\_pitcher\_base       & 88.61 & 75.84                 & 78.80             & 73.53                     \\
021\_bleach\_cleanser    & 88.13 & 75.35                 & 79.02             & 73.60                     \\
024\_bowl                & 87.55 & 74.48                 & 78.33             & 73.20                     \\
025\_mug                 & 88.11 & 75.54                 & 78.70             & 74.40                     \\
035\_power\_drill        & 87.70 & 75.36                 & 78.58             & 73.01                     \\
036\_wood\_block         & 84.60 & 72.02                 & 75.92             & 70.41                     \\
037\_scissors            & 86.45 & 73.63                 & 77.32             & 72.56                     \\
040\_large\_marker       & 88.15 & 75.45                 & 78.34             & 73.93                     \\
051\_large\_clamp        & 74.08 & 63.34                 & 65.89             & 62.80                     \\
052\_extra\_large\_clamp & 67.62 & 57.19                 & 61.03             & 56.01                     \\
061\_foam\_brick         & 87.55 & 74.51                 & 78.40             & 72.72                     \\ \hline
Average                  & 86.22 & 73.67                 & 77.09             & 72.12                     \\
\end{tabular}%
}
\end{table}

\section{Conclusion}
We propose a method that consists of a single feed-forward network that can do the complete inference from data to 6D pose estimation. Our method can be used in real-time taking only $0.12$ seconds to retrieve an accurate 6D pose estimation of a known object present in the scene.
Our method has the best overall performance in both used datasets (LineMOD and YCB-Video). In the LineMOD dataset, we achieved 99.7\% of accuracy, having 0.3\% better accuracy than the second-best method PVN3D. In the YCB-Video dataset, we achieved 98.06\% area under the ADD-S curve which is 1.96\% better than PVN3D.
We performed ablation studies to clarify the impact of the three main architecture components on the overall error rates and found that the removal of these components accounts for a similar error increase on both used datasets, indicating that their benefits are not dependent of the particular data used.

\backmatter

\bmhead{Acknowledgments}

This work was supported by NOVA LINCS (UIDB/04516/2020) with the financial support of FCT-Fundação para a Ciência e a Tecnologia, through national funds, and partially supported by project 026653 (POCI-01-0247-FEDER-026653) INDTECH 4.0 – New technologies for smart manufacturing, cofinanced by the Portugal 2020 Program (PT 2020), Compete 2020 Program and the European Union through the European Regional Development Fund (ERDF).

\bibliography{mybib}


\end{document}